\documentclass[journal,11pt,onecolumn]{IEEEtran}

\usepackage[letterpaper,
            left=1.25in,
            right=1.25in,
            top=1in,
            bottom=1in]{geometry}
\usepackage{setspace}
\usepackage[english]{babel}

\usepackage{ifpdf}

\usepackage{cite} 
\usepackage{url}
\usepackage{hyperref}

\ifCLASSINFOpdf
    \usepackage[pdftex]{graphicx}
    \graphicspath{{./figures/}}
    \DeclareGraphicsExtensions{.pdf,.jpeg,.png}
\else
    \usepackage[dvips]{graphicx}
    \graphicspath{./figures/}
\fi
\usepackage{color}
\usepackage[table]{xcolor}
\usepackage{pgf, tikz, pgfplots}
\usetikzlibrary{shapes, arrows, automata}
\usetikzlibrary{calc,hobby,decorations}
\usepackage[framemethod=tikz]{mdframed} 
\mdfsetup{%
    linecolor=pennblue,
    linewidth=1pt,
    backgroundcolor=pennblue!5}
\usepackage{float}
    \floatstyle{plain}
    \newfloat{SPMfloat}{t}{spm}
\newenvironment{SPMbox}[2]
    {\begin{SPMfloat}[#1]\begin{mdframed}\onehalfspacing\emph{#2.}}
    {\end{mdframed}\end{SPMfloat}}
\usepackage{float}

\usepackage[cmex10]{amsmath}
\usepackage{amsfonts, amssymb, amsthm}
\usepackage{mathrsfs}
\usepackage[mathscr]{euscript}

\usepackage{array}
\usepackage{enumerate}
\usepackage{multirow}
\usepackage{rotating}
\usepackage{caption}
\usepackage{subcaption}
\usepackage{needspace}

\newcommand{\myparagraph}[1]{\needspace{1\baselineskip}\medskip\noindent {\bf #1}}

\input{auxFiles/mySymbol.sty}
\input{auxFiles/pennColors.sty}
\input{auxFiles/myTikzDefs.sty}
\tikzstyle{node value a} = [ empty node, 
                     fill = blue!90,
                     draw = blue!90,
                     text = white]
\tikzstyle{node value b} = [ empty node, 
                     fill = blue!55,
                     draw = blue!55,
                     text = white]
\tikzstyle{node value c} = [ empty node, 
                     fill = blue!35,
                     draw = blue!35,
                     text = white]
\tikzstyle{node value d} = [ empty node, 
                     fill = blue!15,
                     draw = blue!15,
                     text = white]
\tikzstyle{faded node} = [ empty node, 
                     draw = black!30,
                     text = black!30]

\def\Tr{\mathsf{T}}

\def\figStability{3}

\newtheorem{theorem}{\hspace{0pt}\bf Theorem}

\newtheorem{definition}{\hspace{0pt}\bf Definition}

\begin{document}

\title{Graphs, Convolutions, and Neural Networks\\{\vspace{-0.5cm}\Large From Graph Filters to Graph Neural Networks}}

\author{Fer\hspace{0.015cm}nando~Gama,~
        Elvin~Isufi,~
        Geert~Leus,~
        and~Alejandro~Ribeiro
	\thanks{Work in this paper is supported by NSF CCF 1717120, ARO W911NF1710438, ARL DCIST CRA W911NF-17-2-0181, ISTC-WAS and Intel DevCloud. F. Gama, and A. Ribeiro are with the Dept. of Electrical and Systems Eng., Univ. of Pennsylvania, USA. E. Isufi is with the Multimedia Computing Group and G. Leus is with the Circuits and Systems Group, Delft Univ. of Technology, The Netherlands. E-mails: \{fgama, aribeiro\}@seas.upenn.edu, \{e.isufi-1, g.j.t.leus\}@tudelft.nl.}
}

\markboth{IEEE SIGNAL PROCESSING MAGAZINE (ACCEPTED)}%
{Graphs, Convolutions, and Neural Networks}

\maketitle

\begin{abstract}
	Network data can be conveniently modeled as a graph signal, where data values are assigned to nodes of a graph that describes the underlying network topology. Successful learning from network data is built upon methods that effectively exploit this graph structure. In this work, we leverage graph signal processing to characterize the representation space of graph neural networks (GNNs). We discuss the role of graph convolutional filters in GNNs and show that any architecture built with such filters has the fundamental properties of permutation equivariance and stability to changes in the topology. These two properties offer insight about the workings of GNNs and help explain their scalability and transferability properties which, coupled with their local and distributed nature, make GNNs powerful tools for learning in physical networks. We also introduce GNN extensions using edge-varying and autoregressive moving average graph filters and discuss their properties. Finally, we study the use of GNNs in recommender systems and learning decentralized controllers for robot swarms.
\end{abstract}


%
\IEEEpeerreviewmaketitle


\section{Introduction} \label{sec:intro}



Data generated by networks are increasingly common in power grids, robotics, biological, social and economic networks, and recommender systems among others. The irregular and complex nature of these data poses unique challenges so that successful learning is possible only by incorporating the structure into the inner-working mechanisms of the model \cite{Ortega18-GSP}.

Convolutional neural networks (CNNs) have epitomized the success of leveraging the data structure in temporal series and images transforming the landscape of machine learning in the last decade \cite{Goodfellow16-DeepLearning}. CNNs exploit temporal or spatial convolutions to learn an effective nonlinear mapping, scale to large settings, and avoid overfitting \cite[Chapter 10]{Goodfellow16-DeepLearning}. CNNs offer also some degree of mathematical tractability, allowing to derive theoretical performance bounds under domain perturbations \cite{Mallat12-Scattering}. However, convolutions can only be applied to data in regular domains, hence making CNNs ineffective models when learning from irregular network data.

Graphs are used as a mathematical description of network topologies, while the data can be seen as a signal on top of this graph. In recommender systems, for instance, users can be modeled as nodes, their similarities as edges, and the ratings given to items as graph signals. Processing such data by accounting also for the underlying network structure has been the goal of the field of \emph{graph signal processing} (GSP) \cite{Ortega18-GSP}. GSP has extended the concepts of Fourier transform, graph convolutions, and graph filtering to process signals while accounting for the underlying topology.

\emph{Graph convolutional} neural networks (GCNNs) build upon graph convolutions to efficiently incorporate the graph structure into the learning process \cite{Bronstein17-GeometricDeepLearning}. GCNNs consist of a concatenation of layers, in which each layer applies a \emph{graph convolution} followed by a pointwise nonlinearity \cite{Bruna14-DeepSpectralNetworks, Defferrard17-CNNGraphs, Gama19-Architectures, Kipf17-ClassifGCN, Wu19-SGC, Xu19-GIN, Atwood16-Diffusion}. GCNNs exhibit the key properties of permutation equivariance and stability to perturbations \cite{Gama19-Stability, ZouLerman19-Scattering}. The former means GCNNs exploit topological symmetries in the underlying graph, while the latter implies the output is robust to small changes in the graph structure. These results allow GCNNs to scale to large graphs and transfer to different (but similar) scenarios.

Graph convolutions can be exactly modeled by finite impulse response (FIR) graph filters \cite{Ortega18-GSP}. FIR graph filters often require large orders to yield highly discriminatory models, demanding more parameters and an increased computational cost. These limitations are well-understood in the field of GSP and alternative graph filters such as the autoregressive moving average (ARMA) and edge varying graph filters have been proposed to address this \cite{Isufi17-ARMA, Coutino19-Distributed}. ARMA graph filters maintain the convolutional structure but can achieve a similar response with fewer parameters. Contrarily, the edge varying graph filters are inspired by their time varying counterparts and adapt their structure to the specific graph location. The enhanced flexibility of edge varying graph filters requires more parameters but their use lays the foundation of a unified framework for all \emph{graph neural networks} (GNNs) \cite{Isufi20-EdgeNets}, generalizing GCNNs by using non-convolutional graph filters.

In this work, we characterize the representation space of GNNs, obtaining properties and insights that hold irrespective of the specific implementation or set of parameters obtained from training. We highlight the role of graph filters in such a characterization and exploit GSP concepts to derive the permutation equivariance and stability properties that hold for all GCNNs. Section~\ref{sec:graphConv} formally introduces graph convolutions. Section~\ref{sec:GCNN} presents the GCNN and discusses permutation equivariance (Sec.~\ref{subsec:permutationEquivariance}) and stability to graph perturbations (Sec.~\ref{subsec:stability}). Section~\ref{sec:extensionsGraphFilters} generalizes GCNNs by employing alternative graph filters. Section~\ref{sec:applications} provides two applications namely rating prediction in recommender systems and learning decentralized controllers for flocking a robot swarm. Section~\ref{sec:conclusion} contains the paper conclusions and future research directions.


\section{Graphs and Convolutions} \label{sec:graphConv}



We capture the irregular structure of the data by means of an undirected graph $\ccalG=(\ccalV, \ccalE, \ccalW)$ with node set $\ccalV = \{1,\ldots,N\}$, edge set $\ccalE \subseteq \ccalV \times \ccalV$ and weight function $\ccalW: \ccalE \to \reals_{+}$. The neighborhood of node $i \in \ccalV$ is the set of nodes that share an edge with node $i$ and it is denoted as $\ccalN_{i} = \{j \in \ccalV: (j,i) \in \ccalE\}$. An $N \times N$ real symmetric matrix $\bbS$, known as the \emph{graph shift operator}, is associated to the graph and satisfies $[\bbS]_{ij}=s_{ij}=0$ if $(j,i) \notin \ccalE$ for $j \neq i$, i.e., the shift operator has a zero whenever two nodes are disconnected. Common shift operators include the adjacency, Laplacian, and Markov matrices as well as their normalized counterparts \cite{Ortega18-GSP}.
The data on top of this graph forms a graph signal $\bbx \in \reals^{N}$, where the $i$th entry $[\bbx]_{i}=x_{i}$ is the datum of node $i$. Entries $x_{i}$ and $x_{j}$ are pairwise related to each other if there exists an edge $(i,j) \in \ccalE$. The graph signal $\bbx$ can be \emph{shifted} over the nodes by using $\bbS$ so that the $i$th entry of $\bbS\bbx$ is
\begin{equation} \label{eqn:graphShift}
	\big[ \bbS \bbx \big]_{i} 
		= \sum_{j = 1}^{N} [\bbS]_{ij} [\bbx]_{j} 
		= \sum_{j \in \ccalN_{i}} s_{ij} x_{j}.
\end{equation}
where the last equality holds due to the sparsity of $\bbS$ (locality). The output $\bbS \bbx$ is another graph signal where the value at each node is the linear combination of the values of $\bbx$ at the neighbors.

Equipped with the notion of signal shift, we define the \emph{graph convolution} as a linear shift-and-sum operation. Given a set of parameters $\bbh = [h_{0},\ldots,h_{K}]^{\Tr}$, the graph convolution is
\begin{equation} \label{eqn:graphConv}
	\bbH(\bbS) \bbx = \sum_{k=0}^{K} h_{k} \ \bbS^{k} \bbx.
\end{equation}
Operation \eqref{eqn:graphConv} linearly combines the information contained in different neighborhoods. The $k$-shifted signal $\bbS^{k} \bbx$ contains a summary of the information located in the $k$-hop neighborhood and $h_{k}$ weighs this summary. This is a \emph{local} operation since $\bbS^{k} \bbx = \bbS(\bbS^{k-1} \bbx)$ entails $k$ information exchanges with one-hop neighbors [cf. \eqref{eqn:graphShift}]. The graph convolution \eqref{eqn:graphConv} \emph{filters} a graph signal $\bbx$ with a \emph{FIR graph filter} $\bbH(\bbS)$; thus, we refer to the weights $h_{k}$ as the \emph{filter taps} or \emph{filter weights}.

We can gain additional insight about graph convolutions by analyzing \eqref{eqn:graphConv} in the \emph{graph frequency domain} \cite{Ortega18-GSP}. Consider the eigendecomposition of the shift operator $\bbS = \bbV \bbLambda \bbV^{\Tr}$ with orthogonal eigenvector matrix $\bbV \in \reals^{N \times  N}$ and diagonal eigenvalue matrix $\bbLambda \in \reals^{N \times N}$ ordered as $\lambda_{1} \leq \cdots \leq \lambda_{N}$. The eigenvectors $\bbv_{i}$ conform the \emph{graph frequency basis} of graph $\ccalG$ and can be interpreted as signals representing the \emph{graph oscillating modes}, while the eigenvalues $\lambda_{i}$ can be considered as \emph{graph frequencies}. Any signal $\bbx$ can be expressed in terms of these graph oscillating modes
\begin{equation} \label{eqn:GFT}
	\tbx = \bbV^{\Tr} \bbx.
\end{equation}
Operation~\eqref{eqn:GFT} is known as the \emph{graph Fourier transform} (GFT) of $\bbx$, in which entry $[\tbx]_{i} = \tdx_{i}$ denotes the \emph{Fourier coefficient} associated to graph frequency $\lambda_{i}$ and quantifies the contribution of mode $\bbv_{i}$ to the signal $\bbx$ \cite{Ortega18-GSP}. Computing the GFT of the output signal \eqref{eqn:graphConv} yields
\begin{equation} \label{eqn:graphConvGFT}
	\tby 
		= \bbV^{\Tr} \bby 
		= \sum_{k=0}^{K} h_{k} \bbV^{\Tr} \bbV \ \bbLambda^{k} \ \bbV^{\Tr} \bbx 
		= \sum_{k=0}^{K} h_{k} \bbLambda^{k} \ \tbx
		= \bbH(\bbLambda) \tbx
\end{equation}
where $\bbH(\bbLambda)$ is a diagonal matrix with $i$th diagonal element $h(\lambda_{i})$ for $h:\reals \to \reals$ given by
\begin{equation} \label{eqn:frequencyResponse}
	h(\lambda) = \sum_{k=0}^{K} h_{k} \lambda^{k}.
\end{equation}
The function in \eqref{eqn:frequencyResponse} is the \emph{frequency response} of the graph filter $\bbH(\bbS)$ and it is determined solely by the filter taps $\bbh$. The effect that a filter has on a signal depends on the specific graph through the instantiation of $h(\lambda)$ on the eigenvalues $\lambda_{i}$ of $\bbS$. It affects the $i$th frequency content of $\tby$ as
\begin{equation} \label{eqn:filterOutputFrequency}
	\tdy_{i} = h(\lambda_{i}) \tdx_{i}.
\end{equation}
That is, the graph convolution \eqref{eqn:graphConv} modifies the $i$th frequency content $\tdx_{i}$ of the input signal $\bbx$ according to the filter value $h(\lambda_{i})$ at frequency $\lambda_i$. Notice the graph convolution is a pointwise operator in the graph frequency domain, in analogy to the convolution in time and images.


\section{Graph Convolutional Neural Networks} \label{sec:GCNN}



Learning from graph data requires identifying a \emph{representation map} $\bbPhi(\cdot)$ between the data $\bbx$ and the target representation $\bby$ that leverages the graph structure, $\bby = \bbPhi(\bbx; \bbS)$. The image of $\bbPhi$ is known as the \emph{representation space} and determines the space of all possible representations $\bby$ for a given $\bbS$ and any input $\bbx$. One example of a representation map is the graph convolution $\bbPhi(\bbx;\bbS,\ccalH)=\bbH(\bbS)\bbx$ in \eqref{eqn:graphConv}, where set $\ccalH=\{ \bbh\}$ contains the filter coefficients that characterize its representation space \cite{Puschel08-ASP}. To \emph{learn} this map, we consider a cost function $J(\cdot)$ and a training set $\ccalT = \{ \bbx_1,\ldots, \bbx_{|\ccalT|} \}$ with $|\ccalT|$ samples. The \emph{learned map} is then $\bbPhi(\bbx; \bbS, \ccalH^{\star})$ with
\begin{equation} \label{eqn:filterLearning}
     \ccalH^{\star} = \argmin_{\ccalH} \frac{1}{|\ccalT|} \sum_{ \bbx \in \ccalT} J \big(\bbPhi(\bbx; \bbS, \ccalH) \big).
\end{equation}
Typical cost functions include the mean squared error (MSE) or the L1 loss for regression and cross-entropy loss for classification problems \cite{Goodfellow16-DeepLearning}.
Problem \eqref{eqn:filterLearning} consists of finding the $K+1$ filter taps $\ccalH^\star = \{ \bbh^{\star} \}$ that best fit the training data w.r.t. cost $J(\cdot)$, with $K$ being a design choice (a hyperparameter). However, graph convolutions limit the representation power to linear mappings. We can increase the class of mappings that leverage the graph by nesting convolutions into a nonlinearity. The latter leads to the concept of \emph{graph perceptron}, which is formalized next.

\begin{definition}[Graph perceptron] \label{def:graphPerceptron}
A graph perceptron is a mapping that applies an entrywise nonlinearity $\sigma(\cdot)$ to the output of a graph convolution $\bbH(\bbS)\bbx$, i.e.,
\begin{equation} \label{eqn:graphPerceptron}
    \bbPhi(\bbx; \bbS, \ccalH) = \sigma \big( \bbH(\bbS) \bbx \big),
\end{equation}
where set $\ccalH=\{ \bbh \}$ contains the filter coefficients.
\end{definition}
The graph perceptron generates another graph signal obtained as a graph convolution followed by a nonlinearity (e.g., a rectified linear unit, or ReLU for short, $\sigma(z) = \max\{z,0\}$). As such, the graph perceptron captures nonlinear relationships between the data $\bbx$ and the target representation $\bby$. 
By building then a cascade of $L$ graph perceptrons, we get a \emph{multi-layer graph perceptron}, where at layer $\ell$ we compute
\begin{equation} \label{eqn:GCNN}
    \bbx_{\ell} = \sigma\big( \bbH_{\ell} (\bbS) \bbx_{\ell-1} \big), \quad \ell=1,\ldots,L.
\end{equation}
Differently from \eqref{eqn:graphPerceptron}, a multi-layer graph perceptron allows nonlinear signal mixing between nodes. This can be seen in \eqref{eqn:GCNN} where the input of the perceptron at layer $\ell$ is $\bbx_{\ell -1}$, which is in turn the output of the perceptron at layer $\ell-1$, $\bbx_{\ell-1} = \sigma( \bbH_{\ell-1} (\bbS) \bbx_{\ell-2})$. The cascade form allows, therefore, graph convolutions of nonlinear signal transformations coming from the precedent layer. Unrolling this recursion to all layers, we have that the input to the first layer is the data $\bbx_{0} = \bbx$ and the output of the last layer is the estimate of the target representation $\bbx_{L} = \bbPhi(\bbx; \bbS, \ccalH)$; here, the set $\ccalH = \{\bbh_\ell\}_\ell$ contains the filter taps of the $L$ graph filters in \eqref{eqn:GCNN}.

%
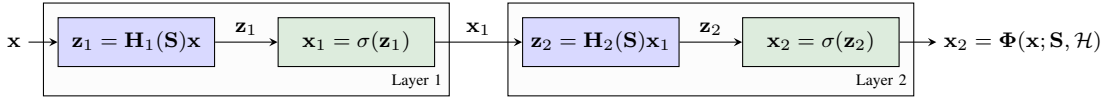
\begin{figure}[t]
    \centering
    {\def \myfactor {0.65} \scriptsize

\def \unit  {\myfactor cm}

\tikzstyle{block} = [ rectangle,
                      minimum width = \unit,
                      minimum height = \unit,
                      fill = blue!15,
                      draw = black,
                      text = black]

\tikzstyle{filter} = [ block,
                      minimum width  = 3.2*\unit,
                      minimum height = 1*\unit]

\tikzstyle{nonlinearity} = [ filter,
                             minimum width  = 3.2*\unit,
                             fill = mygreen!15]

\def \deltainput     {(2.5,0.0)}
\def \deltaoutput    {(0.75,0.0)}
\def \deltalayer     {9.5}
\def \deltaconnector {0.95}
\def \deltasigma     {( 4.5, 0.0)}
\def \deltafeature   {1.5}

\def \one   {$\displaystyle{\bbz_{1}  = \bbH_{1}(\bbS) \bbx    }$}
\def \two   {$\displaystyle{\bbz_{2}  = \bbH_{2}(\bbS) \bbx_{1}}$}
\def \three {$\displaystyle{\bbz_{3}  = \bbH_{3}(\bbS) \bbx_{2}}$}
\def \sigmaone   {$\displaystyle{\bbx_{1} = {\sigma} (\bbz_1)}$}
\def \sigmatwo   {$\displaystyle{\bbx_{2} = {\sigma} (\bbz_2)}$}
\def \sigmathree {$\displaystyle{\bbx_{3} = {\sigma} (\bbz_3)}$}

%
{
\begin{tikzpicture}[scale = \myfactor]

  \pgfdeclarelayer{bg}     
  \pgfsetlayers{bg,main}   

  \node (input) [rectangle, minimum width = 0.1*\unit] {$\bbx$};
  \path (input)      ++ \deltainput node [filter]       (L1 Filter1) {\one};
  \path (L1 Filter1) ++ \deltasigma node [nonlinearity] (L1 F1)      {\sigmaone};
  \path[draw, -stealth] (L1 Filter1.east) -- node [above] {$\bbz_1$} (L1 F1.west);

  \path (L1 Filter1) ++ (\deltalayer,0) node [filter]       (L2 Filter1) {\two};
  \path (L2 Filter1) ++ \deltasigma      node [nonlinearity] (L2 F1)      {\sigmatwo};
  \path[draw, -stealth] (L2 Filter1.east) --  node [above] {$\bbz_2$} (L2 F1.west);

  \path[draw, -stealth] (input.east) -- (L1 Filter1.west);
  \path (L1 F1.east) ++ (0,-\deltaconnector) node [] (aux1) {};
  \path[draw, -stealth] (L1 F1.east) -- node [above] {$\bbx_1$} (L2 Filter1.west);
  \path[draw, -stealth] (L2 F1.east) -- ++ \deltaoutput
                        node [right]{$\bbx_2 = \bbPhi(\bbx; \bbS, \ccalH)$};

  \begin{pgfonlayer}{bg} 
      \path (L1 Filter1.west |- L1 F1.south) ++ (-0.3,-0.6)
           node [filter, anchor = south west,
                 fill = black!1, 
                 minimum width  = 8.3*\unit,
                 minimum height = 1.9*\unit,] 
        (layer)
        {}; 
      \path (layer.south east) ++ (0.0,0.0) node [above left] {\fontsize{6}{6}\selectfont Layer 1};
      \path (L2 Filter1.west |- L2 F1.south) ++ (-0.3,-0.6)
           node [filter, anchor = south west,
                 fill = black!1, 
                 minimum width  = 8.3*\unit,
                 minimum height = 1.9*\unit,] 
        (layer)
        {}; 
      \path (layer.south east) ++ (0.0,0.0) node [above left] {\fontsize{6}{6}\selectfont Layer 2};

  \end{pgfonlayer}

\end{tikzpicture}} 
    }
    \caption{Each blue block represents a linear graph filter and each green block represents a nonlinearity. The concatenation of a convolutional graph filter [cf. \eqref{eqn:graphConv}] and a nonlinearity forms a graph perceptron (Def.~\ref{def:graphPerceptron}) or \emph{layer}. Using a bank of graph convolutional filters [cf. \eqref{eqn:singleFilter}-\eqref{eqn:singleTensorConv}] and cascading several layers, leads to a GCNN. GCNNs are a subset of GNNs, which follow the same structure but consider arbitrary graph filters, see Section~\ref{sec:extensionsGraphFilters}.}
    \label{fig:GNNblockDiagram}
\end{figure}
%

The graph perceptron \eqref{eqn:graphPerceptron} and the multi-layer graph perceptron \eqref{eqn:GCNN} can be viewed as specific \emph{graph convolutional neural networks} (GCNNs). The former is a GCNN of one layer, while the latter is a GCNN of $L$ layers. As it is a good practice in neural networks \cite{Goodfellow16-DeepLearning}, we can substantially increase the representation power of GCNNs by incorporating multiple parallel features per layer. These features are the result of processing multiple input features with a parallel bank of graph filters. Let us consider $F_{\ell-1}$ input graph signal features $\bbx_{\ell-1}^1, \ldots, \bbx_{\ell-1}^{F_{\ell-1}}$ at layer $\ell$. Each input feature $\bbx_{\ell -1}^g$ for $g = 1, \ldots, F_{\ell-1}$ is processed in parallel by $F_\ell$ different graph filters of the form \eqref{eqn:graphConv} to output the $F_\ell$ convolutional features
\begin{equation} \label{eqn:singleFilter}
    \bbu_{\ell}^{fg} = \bbH_{\ell}^{fg}(\bbS) \bbx_{\ell-1}^{g} = \sum_{k=0}^{K} h_{\ell k}^{fg} \bbS^{k} \bbx_{\ell-1}^{g}, \quad f=1,\ldots,F_{\ell}.
\end{equation}
The convolutional features are subsequently summarized along the input index $g$ to yield the aggregated features (see \cite[eq. (13)]{Isufi20-EdgeNets} for a compact matrix-based notation)
\begin{equation} \label{eqn:singleTensorConv}
    \bbu_{\ell}^{f} = \sum_{g=1}^{F_{\ell-1}} \bbH_{\ell}^{fg}(\bbS) \bbx_{\ell-1}^{g}, \quad f=1,\ldots,F_{\ell}.
\end{equation}
The aggregated features are finally passed through a nonlinearity to complete the $\ell$th layer output
\begin{equation} \label{eqn:singleNonlinearity}
    \bbx_{\ell}^{f} = \sigma(\bbu_{\ell}^{f}), \quad f = 1,\ldots,F_{\ell}.
\end{equation}
A GCNN in its complete form\footnote{We omit pooling to emphasize the role of graph filters. Please, refer to \cite{Bruna14-DeepSpectralNetworks, Defferrard17-CNNGraphs, Gama19-Architectures} for pooling methods.} is a concatenation of $L$ layers, in which each layer computes operations \eqref{eqn:singleFilter}-\eqref{eqn:singleTensorConv}-\eqref{eqn:singleNonlinearity}. Differently from the multi-layer graph perceptron GCNN in \eqref{eqn:GCNN}, the complete form employs a parallel bank of $F_{\ell} \times F_{\ell-1}$ graph convolutional filters. This increases the representation power of the mapping and exploits both the stable operation in signal processing, the \emph{convolution}, and the underlying graph structure of the data. The input to the first layer is the data $\bbx_{0}=\bbx$ and the target representation is the collection of $F_{L}$ features of the last layer $[\bbx_{L}^{1},\ldots,\bbx_{L}^{F_{L}}] = \bbPhi(\bbx; \bbS, \ccalH)$, where set $\ccalH = \{\bbh_{\ell}^{fg}\}_{\ell f g}$ collects now the filter taps of all layers. For a given $\bbS$ and fixed hyperparameters $L$, $F_{\ell}$ and $K$, the representation space of a GCNN is characterized by the set of filter coefficients at each layer $\ccalH$. This representation space is different from the one obtained by using linear FIR filters representation maps \cite{Puschel08-ASP}.



\begin{SPMbox}{tp}{Implementations of GCNNs}
Given a matrix representation $\bbS$ and fixed set of hyperparameters (number of layers $L$, filter taps $K$, features $F_{\ell}$ and nonlinearity $\sigma(\cdot)$), the representation space of the GCNN model \eqref{eqn:singleFilter}-\eqref{eqn:singleTensorConv}-\eqref{eqn:singleNonlinearity} is characterized by the set of parameters $\ccalH$ that determine the graph filters. There exist in the literature different implementations for the graph convolution operation \eqref{eqn:singleFilter}, as well as other parametrizations that further restrict this representation space. We overview these in light of the description \eqref{eqn:singleFilter}-\eqref{eqn:singleTensorConv}-\eqref{eqn:singleNonlinearity}.

\medskip \textbf{Same representation space.} Spectral GCNNs \cite{Bruna14-DeepSpectralNetworks} compute \eqref{eqn:singleFilter} in the spectral domain \eqref{eqn:graphConvGFT} and consider the (normalized) Laplacian as the shift $\bbS$; as long as all the eigenvalues of $\bbS$ are different, both \eqref{eqn:graphConvGFT} and \eqref{eqn:singleFilter} are equivalent \cite{Ortega18-GSP}. ChebNets \cite{Defferrard17-CNNGraphs} use a Chebyshev polynomial to compute the graph convolution and consider as $\bbS$ a normalized version of the Laplacian that forces all eigenvalues to be in $[-1,1]$ which is required for the use of Chebyshev polynomials; Chebsyhev polynomials are equivalent to the polynomials in \eqref{eqn:singleFilter}. In summary, we see that \cite{Bruna14-DeepSpectralNetworks, Defferrard17-CNNGraphs} just differ in their implementation of the graph convolution, but all cover the same representation space as the GCNN model \eqref{eqn:singleFilter}-\eqref{eqn:singleTensorConv}-\eqref{eqn:singleNonlinearity} for the specific shift $\bbS$.

\medskip \textbf{Smaller representation space.} GCNs \cite{Kipf17-ClassifGCN} consider \eqref{eqn:singleFilter} with only the one-hop filter tap $h_{\ell 1}^{fg}$ for each layer and each filter, i.e. $K = 1$ and $h_{\ell 0}^{fg} = 0$ for all $\ell$; they adopt a normalized self-looped version of the adjacency as $\bbS$. Simple graph convolutional networks (SGCs) \cite{Wu19-SGC} consider \eqref{eqn:singleFilter} with only the $K$-hop filter tap, i.e. $h_{\ell k}^{fg} = 0$ for all $k<K$; they also adopt a normalized self-looped version of the adjacency as $\bbS$. Graph isomorphism networks (GINs) \cite{Xu19-GIN} consider an order-one polynomial $K=1$ but with $h_{\ell 0}^{fg} = (1+\varepsilon_{\ell})h_{\ell1}^{fg}$ for some pre-defined $\varepsilon_{\ell}$; it adopts the binary adjacency as $\bbS$ and suggests the inclusion of layers with $K=0$ in between layers with $K=1$. Diffusion CNNs \cite{Atwood16-Diffusion} consider a single layer with $F_{1} = N F_{0}$ and the same $K$ filter taps for all input features $h_{1 k}^{fg} = h_{1 k}^{f}$; it adopts the adjacency matrix as $\bbS$. It follows that the representations space of \cite{Kipf17-ClassifGCN, Wu19-SGC, Xu19-GIN, Atwood16-Diffusion} is just a subspace of the representation space of the GCNN model in \eqref{eqn:singleFilter}-\eqref{eqn:singleTensorConv}-\eqref{eqn:singleNonlinearity}.

\medskip

We note that, while the representation space of \cite{Bruna14-DeepSpectralNetworks, Defferrard17-CNNGraphs} is the same as in the GCNN model \eqref{eqn:singleFilter}-\eqref{eqn:singleTensorConv}-\eqref{eqn:singleNonlinearity}, their difference in the implementation of the graph convolution impacts how the optimization space is navigated during training, arriving at different solutions. No particular implementation, however, has consistently outperformed the rest across a wide range of problems. In any case, since the representation space is the same, the characterizations, properties and insights established here apply to all of these. Implementations \cite{Kipf17-ClassifGCN, Wu19-SGC, Xu19-GIN, Atwood16-Diffusion}, on the other hand, further regularize the graph convolution, constraining the representation space to be a subspace of that in the GCNN model. These might be useful in problems with smaller datasets, or where further information on the data structure is available.
\end{SPMbox}

We can \emph{learn} the filter taps by solving problem \eqref{eqn:filterLearning} with the GCNN map $\bbPhi(\bbx; \bbS, \ccalH)$. To do so, we use some optimization method based on stochastic gradient descent \cite{Kingma15-ADAM} and, noting the GCNN is a compositional layered architecture, we also use backpropagation to compute the derivatives of the loss function $J(\cdot)$ with respect to the filter taps $\ccalH$ \cite{Rumelhart86-BackProp}. Since the training data comes from a distribution that has a graph structure $\bbS$, we expect the learned map $\bbPhi(\bbx;\bbS, \ccalH^{\star})$ to \emph{generalize} and perform well for data $\bbx \notin \ccalT$ that come from a similar distribution leveraging $\bbS$. The rationale behind this expectation is that the GCNN is a nonlinear processing architecture that exploits the knowledge the graph carries about the data. Another advantage of a GCNN is its local implementation due to the use of graph convolutions [cf. \eqref{eqn:graphConv}] and pointwise nonlinearities. In fact, all the $F_\ell \times F_{\ell-1}$ convolutional features in \eqref{eqn:singleFilter} are local over the graph as they simply comprise a parallel bank of graph convolutional filters, each of which is local [cf. \eqref{eqn:graphConv}]. Further, since the aggregation step in \eqref{eqn:singleTensorConv} happens across features of the same node and the nonlinearity in \eqref{eqn:singleNonlinearity} is pointwise, these operations are also local and distributable. This built-in characteristic of GCNNs naturally leads to learning solutions that are distributed on the underlying graph.


\subsection{Permutation equivariance} \label{subsec:permutationEquivariance}

A graph shift operator $\bbS$ fixes an arbitrary ordering of the nodes in the graph. Since nodes are naturally unordered, we want the GCNN output to be unaffected by it. That is, we want any change in node ordering to be reflected with the corresponding reordering in the GCNN output.
It turns out GCNNs are unaffected by node labeling --a property known as permutation equivariance-- as stated by the following theorem.

%
\begin{theorem}[Permutation Equivariance \cite{Gama19-Stability, ZouLerman19-Scattering}] \label{thm:permutationEquivariance}
    Consider an $N \times N$ permutation matrix $\bbP$ and the permutations of the shift operator $\hbS = \bbP^{\Tr} \bbS \bbP$ and of the input data $\hbx = \bbP^{\Tr} \bbx$.
    %
    For a GCNN $\bbPhi(\cdot)$, it holds that
    \begin{equation} \label{eqn:permutationEquivariance}
        \bbPhi(\hbx; \hbS, \ccalH) = \bbP^{\Tr} \bbPhi(\bbx; \bbS, \ccalH).
    \end{equation}
\end{theorem}
%

%
\begin{figure*}[t]
    \centering
    \begin{subfigure}{0.25\textwidth}
        \centering
        {\def \myfactor {0.43}
%


{\fontsize{5}{5}\selectfont\begin{tikzpicture}[scale = \myfactor]

  \node                         []     (center) {};
  \path (center) ++ (  0:1.4) node [empty node]   (1) {$1$}  ++ ( 0.5,  0.5) node {$x_1$};
  \path (center) ++ ( 60:1.4) node [node value a] (2) {$2$}  ++ ( 0.6,  0.4) node {$x_2$};
  \path (center) ++ (120:1.4) node [empty node]     (3) {$3$}  ++ (-0.6,  0.4) node {$x_3$};
  \path (center) ++ (180:1.4) node [node value a] (4) {$4$}  ++ (-0.5,  0.5) node {$x_4$};
  \path (center) ++ (240:1.4) node [empty node]   (5) {$5$}  ++ (-0.6, -0.4) node {$x_5$};
  \path (center) ++ (300:1.4) node [empty node]   (6) {$6$}  ++ ( 0.6, -0.4) node {$x_6$};
  \path (center) ++ (  0:3.0) node [empty node]   (7)  {$7$}  ++ ( 0.5,  0.5) node {$x_7$};
  \path (center) ++ ( 60:3.0) node [node value c] (8)  {$8$}  ++ ( 0.5,  0.5) node {$x_8$};
  \path (center) ++ (120:3.0) node [node value a] (9)  {$9$}  ++ ( 0.5,  0.5) node {$x_9$};
  \path (center) ++ (180:3.0) node [node value c] (10) {$10$} ++ (-0.5,  0.5) node {$x_{10}$};
  \path (center) ++ (240:3.0) node [empty node]   (11) {$11$} ++ (-0.5, -0.5) node {$x_{11}$};
  \path (center) ++ (300:3.0) node [empty node]   (12) {$12$} ++ ( 0.9, -0.5) node {$x_{12}$};

  \path (1)  edge [tight edge] node {} (2);
  \path (2)  edge [tight edge] node {} (3);
  \path (3)  edge [tight edge] node {} (4);
  \path (4)  edge [tight edge] node {} (5);
  \path (5)  edge [tight edge] node {} (6);
  \path (6)  edge [tight edge] node {} (1);

  \path (7)   edge [tight edge] node {}  (8);
  \path (8)   edge [tight edge] node {}  (9);
  \path (9)   edge [tight edge] node {} (10);
  \path (10)  edge [tight edge] node {} (11);
  \path (11)  edge [tight edge] node {} (12);
  \path (12)  edge [tight edge] node {}  (7);

  \path (1)  edge [tight edge] node {}  (7);
  \path (2)  edge [tight edge] node {}  (8);
  \path (3)  edge [tight edge] node {}  (9);
  \path (4)  edge [tight edge] node {} (10);
  \path (5)  edge [tight edge] node {} (11);
  \path (6)  edge [tight edge] node {} (12);

\end{tikzpicture}} 
        }
        \vspace{-0.2cm}
        \caption{Graph $\bbS$ and signal $\bbx$}
        \label{original}
    \end{subfigure}
    \hspace{0.35cm}
    \begin{subfigure}{0.25\textwidth}
        \centering
        {\def \myfactor {0.43}
%


{\fontsize{5}{5}\selectfont\begin{tikzpicture}[scale = \myfactor]

  \node                         []     (center) {};
  \path (center) ++ (  0:1.4) node [node value c] (1) {$1$}  ++ ( 0.5,  0.5) node {$x_1$};
  \path (center) ++ ( 60:1.4) node [empty node]   (2) {$2$}  ++ ( 0.6,  0.4) node {$x_2$};
  \path (center) ++ (120:1.4) node [empty node]   (3) {$3$}  ++ (-0.6,  0.4) node {$x_3$};
  \path (center) ++ (180:1.4) node [empty node]   (4) {$4$}  ++ (-0.5,  0.5) node {$x_4$};
  \path (center) ++ (240:1.4) node [node value c] (5) {$5$}  ++ (-0.6, -0.4) node {$x_5$};
  \path (center) ++ (300:1.4) node [node value a]   (6) {$6$}  ++ ( 0.6, -0.4) node {$x_6$};
  \path (center) ++ (  0:3.0) node [node value a] (7)  {$7$}  ++ ( 0.5,  0.5) node {$x_7$};
  \path (center) ++ ( 60:3.0) node [empty node]   (8)  {$8$}  ++ ( 0.5,  0.5) node {$x_8$};
  \path (center) ++ (120:3.0) node [empty node]   (9)  {$9$}  ++ ( 0.5,  0.5) node {$x_9$};
  \path (center) ++ (180:3.0) node [empty node]   (10) {$10$} ++ (-0.5,  0.5) node {$x_{10}$};
  \path (center) ++ (240:3.0) node [node value a] (11) {$11$} ++ (-0.5, -0.5) node {$x_{11}$};
  \path (center) ++ (300:3.0) node [empty node] (12) {$12$} ++ ( 0.9, -0.5) node {$x_{12}$};

  \path (1)  edge [tight edge] node {} (2);
  \path (2)  edge [tight edge] node {} (3);
  \path (3)  edge [tight edge] node {} (4);
  \path (4)  edge [tight edge] node {} (5);
  \path (5)  edge [tight edge] node {} (6);
  \path (6)  edge [tight edge] node {} (1);

  \path (7)   edge [tight edge] node {}  (8);
  \path (8)   edge [tight edge] node {}  (9);
  \path (9)   edge [tight edge] node {} (10);
  \path (10)  edge [tight edge] node {} (11);
  \path (11)  edge [tight edge] node {} (12);
  \path (12)  edge [tight edge] node {}  (7);

  \path (1)  edge [tight edge] node {}  (7);
  \path (2)  edge [tight edge] node {}  (8);
  \path (3)  edge [tight edge] node {}  (9);
  \path (4)  edge [tight edge] node {} (10);
  \path (5)  edge [tight edge] node {} (11);
  \path (6)  edge [tight edge] node {} (12);

\end{tikzpicture}}
        }
        \vspace{-0.2cm}
        \caption{Graph $\bbS$ and signal $\hbx$}
        \label{symmetries}
    \end{subfigure}
    \hspace{0.35cm}
    \begin{subfigure}{0.25\textwidth}
        \centering
        {\def \myfactor {0.43}
%


{\fontsize{5}{5}\selectfont\begin{tikzpicture}[scale = \myfactor]

  \node                         []     (center) {};
  \path (center) ++ (  0:1.4) node [empty node]   (1) {$10$}  ++ ( 0.5,  0.5) node {$x_1$};
  \path (center) ++ ( 60:1.4) node [node value a] (2) {$11$}  ++ ( 0.6,  0.4) node {$x_2$};
  \path (center) ++ (120:1.4) node [empty node]     (3) {$12$}  ++ (-0.6,  0.4) node {$x_3$};
  \path (center) ++ (180:1.4) node [node value a] (4) {$7$}  ++ (-0.5,  0.5) node {$x_4$};
  \path (center) ++ (240:1.4) node [empty node]   (5) {$8$}  ++ (-0.6, -0.4) node {$x_5$};
  \path (center) ++ (300:1.4) node [empty node]   (6) {$9$}  ++ ( 0.6, -0.4) node {$x_6$};
  \path (center) ++ (  0:3.0) node [empty node]   (7)  {$4$}  ++ ( 0.5,  0.5) node {$x_7$};
  \path (center) ++ ( 60:3.0) node [node value c] (8)  {$5$}  ++ ( 0.5,  0.5) node {$x_8$};
  \path (center) ++ (120:3.0) node [node value a] (9)  {$6$}  ++ ( 0.5,  0.5) node {$x_9$};
  \path (center) ++ (180:3.0) node [node value c] (10) {$1$} ++ (-0.5,  0.5) node {$x_{10}$};
  \path (center) ++ (240:3.0) node [empty node]   (11) {$2$} ++ (-0.5, -0.5) node {$x_{11}$};
  \path (center) ++ (300:3.0) node [empty node]   (12) {$3$} ++ ( 0.9, -0.5) node {$x_{12}$};

  \path (1)  edge [tight edge] node {} (2);
  \path (2)  edge [tight edge] node {} (3);
  \path (3)  edge [tight edge] node {} (4);
  \path (4)  edge [tight edge] node {} (5);
  \path (5)  edge [tight edge] node {} (6);
  \path (6)  edge [tight edge] node {} (1);

  \path (7)   edge [tight edge] node {}  (8);
  \path (8)   edge [tight edge] node {}  (9);
  \path (9)   edge [tight edge] node {} (10);
  \path (10)  edge [tight edge] node {} (11);
  \path (11)  edge [tight edge] node {} (12);
  \path (12)  edge [tight edge] node {}  (7);

  \path (1)  edge [tight edge] node {}  (7);
  \path (2)  edge [tight edge] node {}  (8);
  \path (3)  edge [tight edge] node {}  (9);
  \path (4)  edge [tight edge] node {} (10);
  \path (5)  edge [tight edge] node {} (11);
  \path (6)  edge [tight edge] node {} (12);

\end{tikzpicture}} 
        }
        \vspace{-0.2cm}
        \caption{Graph $\hbS$ and signal $\hbx$}
        \label{reordering}
    \end{subfigure}
    \caption{Permutation equivariance of GCNNs. The output of a GCNN is equivariant to graph permutations (Theorem \ref{thm:permutationEquivariance}). This means independence from labeling and shows GCNNs exploit internal signal symmetries. Signals in \subref{original} and \subref{symmetries} are different on the same graph but they are permutations of each other --interchange inner and outer hexagons and rotate $180^{\circ}$ [c.f. \subref{reordering}]. A GCNN would learn how to classify the signal in \subref{symmetries} from seeing examples of the signal in \subref{original}. Integers represent labels, while colors signal values.}
    \label{fig:permutationEquivariance}
\end{figure*}
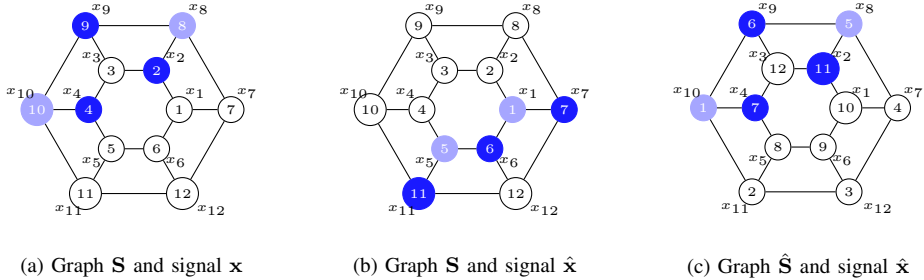
%

Theorem~\ref{thm:permutationEquivariance} states that a node reordering results in a corresponding reordering of the GCNN output, implying GCNNs are independent of node labeling. Theorem~\ref{thm:permutationEquivariance} implies also that graph convolutions exploit the inherent symmetries present in a graph to improve data processing. If the graph exhibits several nodes with the same topological neighborhood (graph symmetries), then learning how to process data in any of these nodes can be translated to every other node with the same topological neighborhood. This allows GCNNs to learn from fewer samples and generalize easier to signals located at any topologically similar neighborhood, see Fig.~\ref{fig:permutationEquivariance}.


\subsection{Stability to perturbations} \label{subsec:stability}

Since real graphs rarely exhibit perfect symmetries, we are interested in more general changes to the underlying graph support than just permutations. For instance, in problems where the graph $\bbS$ is fixed but unknown, we need to use an estimate $\hbS$ of it but we still want the GCNN to work well as long as the estimate is good (Section~\ref{subsec:recSys}). On another set of problems, the graph support may naturally differ from training $\bbS$ to testing $\hbS$, a scenario known as \emph{transfer learning} (Section~\ref{subsec:flocking}). In these cases, we need the GCNN to have a similar performance whether they run on $\bbS$ or on $\hbS$ as long as both graphs are similar. To measure the similarity between graphs $\bbS$ and $\hbS$, and in light of Theorem~\ref{thm:permutationEquivariance}, we define next the relative distance modulo permutation.

%
\begin{definition}[Relative distance] \label{def:relativeDistance}
    Consider the set of all permutation matrices $\ccalP = \{\bbP \in \{0,1\}^{N \times N} : \bbP^{\Tr} \bbone = \bbone, \bbP \bbone = \bbone \}$. For two graphs $\bbS$ and $\hbS$ with the same number of nodes, we define the set of \emph{relative error matrices} as
    \begin{equation} \label{eqn:relativeErrorSet}
        \ccalR(\bbS, \hbS) = \{ \bbE : \bbP^{\Tr} \hbS \bbP = \bbS + (\bbE \bbS + \bbS \bbE) \ , \ \bbP \in \ccalP \}.
    \end{equation}
    The \emph{relative distance} modulo permutations between $\bbS$ and $\hbS$ is then defined as
    \begin{equation} \label{eqn:relativeDistance}
        d(\bbS, \hbS) = \min_{\bbE \in \ccalR(\bbS, \hbS)} \| \bbE \|
    \end{equation}
    where $\| \cdot \|$ indicates the operator norm. We denote by $\bbE^{\star}$ and $\bbP^{\star}$ the relative error matrix and the permutation matrix that minimize \eqref{eqn:relativeDistance}, respectively.
\end{definition}
%
\noindent We readily see that if $d(\bbS, \hbS) = 0$, then $\hbS$ is a permutation of $\bbS$, and thus the relative distance \eqref{eqn:relativeDistance} measures how far $\bbS$ and $\hbS$ are from being permutations of each other. We note that, unlike the absolute perturbation model, the relative distance (Def.~\ref{def:relativeDistance}) accurately reflects changes to both the edge weights and the topology structure \cite{Gama19-Stability}.

The change in the output of a GCNN due to a change in the underlying support is bounded for GCNNs whose constitutive graph filters are integral Lipschitz.

%
\begin{definition}[Integral Lipschitz filters] \label{def:integralLipschitz}
    We say a filter $\bbH(\bbS)$ is \emph{integral Lipschitz} if its frequency response $h(\lambda)$ [cf. \eqref{eqn:frequencyResponse}] is such that $|h(\lambda)| \leq 1$ and its derivative $h'(\lambda)$ satisfies $|\lambda h'(\lambda)| \leq C$ for some finite constant $C$.
\end{definition}
%
\noindent The derivative condition $|\lambda h'(\lambda)| \leq C$ implies integral Lipschitz filters have frequency responses that can vary rapidly around $\lambda = 0$ but are flat for $\lambda \to \infty$, see Fig.~\figStability a. GCNNs that use integral Lipschitz filters are stable under relative perturbations. This means the change in the output due to changes in the underlying graph is bounded by the size of the perturbation [cf. Def.~\ref{def:relativeDistance}].

%
\begin{theorem}[Stability \cite{Gama19-Stability}] \label{thm:stability}
    Let $\bbS$ and $\hbS$ be two different graphs with the same number of nodes such that their relative distance is $d(\bbS, \hbS) \leq \varepsilon$ [cf. Def.~\ref{def:relativeDistance}]. Let $\bbPhi(\cdot; \cdot, \ccalH)$ be a multi-layer graph perceptron GCNN [cf. \eqref{eqn:GCNN}] where all filters $\ccalH$ are integral Lipschitz with constant $C$ [cf. Def.~\ref{def:integralLipschitz}]. Then, it holds that
    \begin{equation} \label{eqn:stability}
        \| \bbPhi(\bbx; \bbS, \ccalH) - \bbPhi(\bbP^{\star^{\Tr}} \bbx; \bbP^{\star^{\Tr}} \hbS \bbP^{\star}, \ccalH)\|
            \leq 2C \ (1+ \delta\sqrt{N}) \ L \ \varepsilon \ \| \bbx \| + \ccalO(\varepsilon^{2})
    \end{equation}
    where $\delta = (\| \bbU - \bbV\| + 1)^{2} - 1$ is the eigenvector misalignment between the eigenbasis $\bbV$ of $\bbS$ and the eigenbasis $\bbU$ of the relative error matrix $\bbE^{\star}$, with $\bbE^{\star}$ and $\bbP^{\star}$ given in Definition \ref{def:relativeDistance}.
\end{theorem}
%

\noindent Theorem~\ref{thm:stability} proves that a change $\varepsilon$ in the shift operator causes a change proportional to $\varepsilon$ in the GCNN output. The proportionality constant has the term $C$ that depends on the filter design, and the term $(1+\delta\sqrt{N})$ that depends on the specific perturbation. But it also has a constant factor $L$ that depends on the depth of the architecture implying deeper GCNNs are less stable.



\begin{SPMbox}{tp}{Insights on stability}
To offer further insight into Theorem~\ref{thm:stability}, consider the particular case where the perturbation $\hbS$ is an edge dilation of a graph $\bbS$, i.e. $\hbS = (1+\varepsilon) \bbS$, where all edges are increased proportionally by a factor $\varepsilon$. The relative error matrix is $\bbE = (\varepsilon/2) \bbI$ so that the relative distance is $d(\bbS, \hbS) = \|\bbE\| \leq \varepsilon$. The graph dilation changes the eigenvalues to $\hat{\lambda}_{i} = (1+\varepsilon) \lambda_{i}$ while the eigenvectors remain the same. We note that, even if $\varepsilon$ is small, the change in eigenvalues could be large if $\lambda_{i}$ is large, see Fig.~\figStability a.

{
\bigskip
\def \thisplotscale {1.8}

\begin{tabular}{ccc}
    {\scriptsize 

\def \unit {\thisplotscale cm}

\def \frequencyresponse 
     { 0.9 - 0.7*exp(-(0.7*(x-1.6))^2) }

\hspace{-2.9mm}
\begin{tikzpicture}[x = 1*\unit, y=1*\unit]

\def \factorx {2.4/8}
\def \deltax  {0.5*\factorx}
\def \shadeshift  {0.05}

\path [fill=black!20, opacity = 0.5] 
              (\deltax - 0.001*\factorx - \shadeshift, 0.00) rectangle 
              (\deltax + 0.030*\factorx + \shadeshift, 1.00);
\path [fill=black!20, opacity = 0.5] 
              (\deltax + 3.393*\factorx - \shadeshift, 0.00) rectangle 
              (\deltax + 3.770*\factorx + \shadeshift, 1.00);
\path [fill=black!20, opacity = 0.5] 
              (\deltax + 6.048*\factorx - \shadeshift, 0.00) rectangle 
              (\deltax + 6.720*\factorx + \shadeshift, 1.00);

\begin{axis}[scale only axis,
             width  = 2.4*\unit,
             height = 1*\unit,
             xmin = -0.5, xmax=7.5,
             xtick = {0.03, -0.01, 3.77, 3.393, 6.72, 6.048},
             xticklabels = {\red{$\qquad\hat{\lambda}_1\phantom{\lambda}$},
                            \blue{$\lambda_1\ \ $}, 
                            \red{$\quad\hat{\lambda}_i\phantom{\lambda}$}, 
                            \blue{$\lambda_i$},
                            \red{$\quad\hat{\lambda}_{N}\phantom{\lambda}$},
                            \blue{$\lambda_N$}},
             ymin = -0, ymax = 1.15,
             ytick = {-1},
             typeset ticklabels with strut,
             enlarge x limits=false]

\addplot+[samples at = {0.03, 0.91, 1.57, 
                        2.63, 3.77, 4.51, 
                        5.60, 6.72}, 
          color = red!60, 
          ycomb, 
          mark=otimes*, 
          mark options={red!60}]
         {\frequencyresponse};

\addplot+[samples at = {-0.01, 0.819, 1.413, 
                        2.367, 3.393, 4.059, 
                        5.04, 6.048}, 
          color = blue!60, 
          ycomb, 
          mark=oplus*, 
          mark options={blue!60}]
         {\frequencyresponse};

\addplot[ domain=-0.5:7.5, 
          samples = 80, 
          color = black,
          line width = 1.2]
         {\frequencyresponse};

\end{axis}
\end{tikzpicture}

}
    &
    \hspace{-0.5cm}
    {\scriptsize

\def \unit {\thisplotscale cm}

\def \frequencyresponse 
     {1.1*exp(-(2.5*(x-6.048))^2}
\def \frequencyresponsetwo 
     {1.1*exp(-(2.5*(x-5.04))^2}

\hspace{-2.9mm}
\begin{tikzpicture}[x = 1*\unit, y=1*\unit]

\def \factorx {2.4/8}
\def \deltax  {0.5*\factorx}
\def \shadeshift  {0.05}

\path [fill=black!20, opacity = 0.5] 
              (\deltax + 6.048*\factorx - \shadeshift, 0.00) rectangle 
              (\deltax + 6.720*\factorx + \shadeshift, 1.00);

\path [fill=black!20, opacity = 0.5] 
              (\deltax + 5.04*\factorx - \shadeshift, 0.00) rectangle 
              (\deltax + 5.60*\factorx + \shadeshift, 1.00);

\begin{axis}[scale only axis,
             width  = 2.4*\unit,
             height = 1*\unit,
             xmin = -0.5, xmax=7.5,
             xtick = {5.60, 5.04, 6.72, 6.048},
             xticklabels = {\red{$\quad\qquad\hat{\lambda}_{N-1}\phantom{\lambda_{N-1}}$},
                            \blue{$\lambda_{N-1}\qquad\quad  $}, 
                            \red{$\quad\qquad\hat{\lambda}_{N}\phantom{\lambda}$},
                            \blue{$\,\qquad\lambda_N$}},
             ymin = -0, ymax = 1.15,
             ytick = {-1},
             typeset ticklabels with strut,
             enlarge x limits=false]

\addplot+[samples at = {0.03, 0.91, 1.57, 
                        2.63, 3.77, 4.51}, 
          color = red!60, 
          ycomb, 
          mark=otimes*, 
          mark options={red!60}]
         {0};

\addplot+[samples at = {6.72, 5.60}, 
          color = red!60, 
          ycomb, 
          mark=otimes*, 
          mark options={red!60}]
         {1};

\addplot+[samples at = {-0.01, 0.819, 1.413, 
                        2.367, 3.393, 4.059}, 
          color = blue!60, 
          ycomb, 
          mark=oplus*, 
          mark options={blue!60}]
         {0};

\addplot+[samples at = {6.048, 5.04}, 
          color = blue!60, 
          ycomb, 
          mark=oplus*, 
          mark options={blue!60}]
         {1};

\addplot[ domain=-0.5:5.5, 
          samples = 2, 
          color = black,
          line width = 1.2]
         {0};

\addplot[ domain=5.0:7.5, 
          samples = 70, 
          color = black,
          line width = 1.2]
         {\frequencyresponse};

\addplot[ domain=4.0:7.0, 
          samples = 70, 
          color = black,
          line width = 1.2]
         {\frequencyresponsetwo};
\addplot[ domain=7.0:7.5, 
          samples = 2, 
          color = black,
          line width = 1.2]
         {0};

\end{axis}
\end{tikzpicture}

}
    &
    \hspace{-1cm}
    {\scriptsize

\def \unit {\thisplotscale cm}

\def \frequencyresponse 
     {   0.8*exp(-(1*(x-1.2))^2) 
       + 0.7*exp(-(0.7*(x-4))^2) 
       + 0.8*exp(-(1.4*(x-6))^2) 
       + 0.1}

\hspace{-2.9mm}
\begin{tikzpicture}[x = 1*\unit, y=1*\unit]

\def \factorx {2.4/8}
\def \deltax  {0.5*\factorx}
\def \shadeshift  {0.05}

\path [fill=black!20, opacity = 0.5] 
              (\deltax - 0.001*\factorx - \shadeshift, 0.00) rectangle 
              (\deltax + 0.030*\factorx + \shadeshift, 1.00);
\path [fill=black!20, opacity = 0.5] 
              (\deltax + 3.393*\factorx - \shadeshift, 0.00) rectangle 
              (\deltax + 3.770*\factorx + \shadeshift, 1.00);
\path [fill=black!20, opacity = 0.5] 
              (\deltax + 6.048*\factorx - \shadeshift, 0.00) rectangle 
              (\deltax + 6.720*\factorx + \shadeshift, 1.00);

\begin{axis}[scale only axis,
             width  = 2.4*\unit,
             height = 1*\unit,
             xmin = -0.5, xmax=7.5,
             xtick = {0.03, -0.01, 3.77, 3.393, 6.72, 6.048},
             xticklabels = {\red{$\qquad\hat{\lambda}_1\phantom{\lambda}$},
                            \blue{$\lambda_1\ \ $}, 
                            \red{$\quad\hat{\lambda}_i\phantom{\lambda}$}, 
                            \blue{$\lambda_i$},
                            \red{$\quad\hat{\lambda}_{N}\phantom{\lambda}$},
                            \blue{$\lambda_N$}},
             ymin = -0, ymax = 1.15,
             ytick = {-1},
             typeset ticklabels with strut,
             enlarge x limits=false]

\addplot+[color = red!60, 
          ycomb, 
          mark=otimes*, 
          mark options={red!60}]
          coordinates { (0.03, 0.20)
                        (0.91, 0.10)
                        (1.57, 0.10)
                        (2.63, 0.07)
                        (3.77, 0.12)
                        (4.51, 0.12)
                        (5.60, 0.21)
                        (6.72, 0.70)};

\addplot+[samples at = {-0.01, 0.819, 1.413, 
                        2.367, 3.393, 4.059, 
                        5.04, 6.048}, 
          color = blue!60, 
          ycomb, 
          mark=oplus*, 
          mark options={blue!60}]
          coordinates { (-0.010, 0.20)
                        ( 0.819, 0.10)
                        ( 1.413, 0.10)
                        ( 2.367, 0.07)
                        ( 3.393, 0.12)
                        ( 4.059, 0.12)
                        ( 5.040, 0.21)
                        ( 6.048, 0.70)};

\end{axis}
\end{tikzpicture}}
    \\
    {\scriptsize (a) Integral Lipschitz filters} & 
    {\scriptsize \hspace{-0.5cm} (b) High eigenvalue features} & 
    {\scriptsize \hspace{-1cm} (c) Frequency mixing} \\
    \multicolumn{3}{p{\textwidth}}{\vspace{-0.5cm}\begin{singlespace}{\footnotesize Figure \figStability. (a) Frequency response for an integral Lipschitz filter (in black), eigenvalues for $\bbS$ (in blue) and eigenvalues for $\hbS$ (in red). Larger eigenvalues exhibit a larger change. (b) Separating energy located at $\lambda_{N-1}$ from that at $\lambda_{N}$ requires filters with sharp transitions that are not integral Lipschitz. Then, a change in eigenvalues renders these filters useless (they are not stable). (c) Applying a ReLU to a signal with all its energy located at $\lambda_{N}$ results in a signal with energy spread through the spectrum. Information on low eigenvalues can be discriminated in a stable fashion.\vspace{-0.2cm}}\end{singlespace}}\vspace{-0.5cm}
\end{tabular}
\stepcounter{figure}
}

\medskip This observation that even small perturbations lead to large changes in the eigenvalues can considerably affect the output of a graph filter causing instability, unless the graph filters are carefully designed. To see this, consider first the output of a graph filter in the frequency domain, $\tdy_{i} = h(\lambda_{i}) \tdx_{i}$ [cf. \eqref{eqn:filterOutputFrequency}]. With the graph dilation, the frequency response gets instantiated at $\hat{\lambda}_{i} = (1+\varepsilon) \lambda_{i}$ instead of $\lambda_{i}$, so the $i$th frequency content is now $\hat{\tdy}_{i} = h(\hat{\lambda}_{i}) \tdx_{i}$. The change between the original $i$th frequency content of the output $\tdy_{i}$ and the perturbed one $\hat{\tdy}_{i}$ depends on how much $h(\lambda_{i})$ changes with respect to $h(\hat{\lambda}_{i})$, and thus can be quite large for large $\lambda$. So if we want $\tdy_{i}$ to be close to $\hat{\tdy}_{i}$ for stability, we need to have frequency responses $h(\lambda)$ that have a flat response for large eigenvalues, see Fig.~\figStability a. Integral Lipschitz filters do have a flat response at large eigenvalues and thus are stable.

\medskip The cost to pay for stability is that integral Lipschitz filters cannot discriminate information located at higher eigenvalues. As seen in Fig.~\figStability b, discriminative filters are narrow filters. Then, if even a small perturbation causes a large change in the instantiated eigenvalue (as is the case for large eigenvalues), the filter output changes to a zero output, and thus is not stable. Thus, linear graph filters exhibit a trade-off between discriminability and stability; a trait shared by regular convolutional filters \cite{Mallat12-Scattering}.

\medskip GCNNs incorporate pointwise nonlinearities in the graph perceptron. This nonlinear operation has a frequency mixing effect (akin to demodulation) by which the signal energy is spilled throughout the spectrum, see Fig.~\figStability c. Thus, energy from large eigenvalues now appears in smaller eigenvalues. This new low-eigenvalue frequency content can be captured and discriminated by subsequent filters in a stable manner. Therefore, pointwise nonlinearities make GCNNs information processing architectures that are both stable and selective.
\end{SPMbox}


\section{Extensions: General Graph Filters} \label{sec:extensionsGraphFilters}



Oftentimes, the GCNN would require highly sharp filter responses to discriminate between classes. We can increase the discriminatory power by either increasing the filter order $K$ or changing the filter type $\bbH(\bbS)$ in the graph perceptron \eqref{eqn:graphPerceptron}. Increasing $K$ is not always feasible as it leads to more filter coefficients, a higher complexity, and numerical issues related to the higher order powers of the shift operator $\bbS^k$. Instead, changing the filter type allows implementing another family of graph neural networks (GNNs) with different properties. We present two alternative filters that provide different insights on how to design more general GNNs: \emph{the autoregressive moving average} (ARMA) graph filter \cite{Isufi17-ARMA} and \emph{the edge varying} graph filter \cite{Coutino19-Distributed}.

\subsection{ARMANet}\label{subsec:ARMANet}

An ARMA graph filter operates also pointwise in the spectral domain $\tdy_{i} = h(\lambda_{i}) \tdx_{i}$ [cf. \eqref{eqn:filterOutputFrequency}] but it is characterized by the rational frequency response
\begin{equation}\label{eq.ARMAResp}
h(\lambda) = \frac{\sum_{q = 0}^Qb_q\lambda^q}{1 + \sum_{p = 1}^Pa_p\lambda^p}.
\end{equation}
The frequency response is now controlled by $P$ denominator coefficients $\bba = [a_1, \ldots, a_P]^{\Tr}$ and $Q+1$ numerator coefficients $\bbb = [b_0, \ldots, b_Q]^{\Tr}$. The rational frequency responses in \eqref{eq.ARMAResp} span an equivalent space to that of graph filters in \eqref{eqn:graphConv}. However, the spectral equivalence does not imply that the two filters have the same properties. ARMA filters implement rational frequency responses rather than polynomial ones as FIR filters do \eqref{eqn:frequencyResponse}. Therefore, we expect them to achieve a sharper response with lower orders of $P$ and $Q$ such that $P+Q < K$. Replacing the spectral variable $\lambda$ with the shift operator $\bbS$ allows us to write the ARMA output $\bby = \bbH(\bbS)\bbx$ as
\begin{equation}\label{eq.ARMAfilt}
\bby = \Big(\bbI + \sum_{p = 1}^Pa_p\bbS^p	\Big)^{-1}\Big(\sum_{q = 0}^Qb_q\bbS^q\Big)\bbx := \bbP(\bbS)^{-1}\bbQ(\bbS)\bbx
\end{equation}
where $\bbP(\bbS) := \bbI + \sum_{p = 1}^Pa_p\bbS^p$ and $\bbQ := \sum_{q = 0}^Qb_1\bbS^q$ are two FIR filters [cf. \eqref{eqn:graphConv}] that allow writing the ARMA filter as $\bbH(\bbS) = \bbP(\bbS)^{-1}\bbQ(\bbS)$. As it follows from \eqref{eq.ARMAfilt}, we need to apply the matrix inverse $\bbP(\bbS)$ to obtain the ARMA output. This, unless the number of nodes is moderate, is computationally unaffordable; hence, we need an iterative method to approximately apply the inverse. Due to its faster convergence, we choose a parallel structure that consists of first transforming the polynomial ratio in \eqref{eq.ARMAfilt} into its partial fraction decomposition form and subsequently using the Jacobi method to approximately apply the inverse. While also other Krylov approaches are possible to solve \eqref{eq.ARMAfilt}, the parallel Jacobi method offers a better tradeoff between computational complexity, distributed implementation, and convergence.

\myparagraph{Partial fraction decomposition of ARMA filters.} Consider the rational frequency response $h(\lambda)$ in \eqref{eq.ARMAResp} and let $\bbgamma = [\gamma_1, \ldots, \gamma_P]^{\Tr}$ be the $P$ poles, $\bbbeta = [\beta_1, \ldots, \beta_P]^{\Tr}$ the corresponding residuals and $\bbalpha = [\alpha_0, \ldots, \alpha_K]^{\Tr}$ the direct terms. Then, we can write \eqref{eq.ARMAfilt} in the equivalent form
\begin{equation}\label{eq.PFD_ARMA}
\bby  = \sum_{p = 1}^P\beta_p \bigg(\bbS - \gamma_p\bbI\bigg)^{-1}\bbx + \sum_{k = 0}^K\alpha_k\bbS^k\bbx.
\end{equation}
The equivalence of \eqref{eq.PFD_ARMA} and \eqref{eq.ARMAfilt} implies that instead of learning $\bba$ and $\bbb$ in \eqref{eq.ARMAfilt}, we can learn $\bbalpha$, $\bbbeta$, and $\bbgamma$ in \eqref{eq.PFD_ARMA}. 
To avoid the matrix inverses in the single pole filters, we can approximate each output $\bbu_p$ through the Jacobi method.

\myparagraph{Jacobi method for single pole filters.} We can write the output of the $p$th single pole filter $\bbu_p$ in the equivalent linear equation form $(\bbS - \gamma_p\bbI)\bbu_p = \beta_p\bbx$. The Jacobi algorithm requires separating $(\bbS - \gamma_p\bbI)$ into its diagonal and off-diagonal terms. Defining $\bbD = \diag(\bbS)$ as the matrix containing the diagonal of the shift operator, we can write the Jacobi approximation $\bbu_{p\tau}$ of the $p$th single pole filter output $\bbu_p$ at iteration $\tau$ by the recursive expression
\begin{equation}\label{eq.JacRec}
\bbu_{p\tau} = \bigg(\bbD - \gamma_p\bbI\bigg)^{-1}\bigg[\beta_p\bbx - \bigg(\bbS - \bbD\bigg)\bbu_{p(\tau-1)}	\bigg],\qquad\text{with}~\bbu_{p0} = \bbx.
\end{equation}
The inverse in \eqref{eq.JacRec} is now element-wise on the diagonal matrix $(\bbD - \gamma_p\bbI)$. This recursion can be unrolled to all its terms to write an explicit relationship between $\bbu_{p\tau}$ and $\bbx$. To do that, we define the parameterized shift operator $\bbR(\gamma_p) = - \big(\bbD - \gamma_p\bbI\big)^{-1}\big(\bbS - \bbD\big)$ and use it to write the $T$th Jacobi recursion as
\begin{equation}\label{eq.JacobiK}
\bbu_{pT} = \beta_p \sum_{\tau = 0}^{T-1}\bbR^\tau(\gamma_p)\bbx + \bbR^T(\gamma_p)\bbx. 
\end{equation}

For a convergent Jacobi method, $\bbu_{pT}$ converges to the single pole output $\bbu_p$. However, in a practical setting we truncate \eqref{eq.JacobiK} for a finite $T$. We can then write the single pole filter output as $\bbu_{pT} := \bbH_T(\bbR(\gamma_{p}))\bbx$, where we define the following FIR filter of order $T$
\begin{equation}\label{eq.FIRparam}
    \bbH_T(\bbR(\gamma_p)) = \beta_p\sum_{\tau = 0}^{T-1}\bbR^\tau(\gamma_p) + \bbR^T(\gamma_p).
\end{equation}
with the parametric shift operator $\bbR(\gamma)$. In other words, a single pole filter is approximated by a graph convolutional filter of the form \eqref{eqn:graphConv} in which the shift operator $\bbS$ is substituted by $\bbR(\gamma)$. This parametric convolutional filter uses coefficients $\beta_p$ for $\tau = 0, \ldots, T-1$ and $1$ for $\tau = T$.

\myparagraph{Jacobi ARMA filters and ARMANets.} Assuming we use truncated Jacobi iterations of order $T$ to approximate all single pole filters in \eqref{eq.PFD_ARMA}, we can write the ARMA filter as
\begin{equation}\label{eq.ARMAtot}
    \bbH(\bbS) = \sum_{p = 1}^P\bbH_T(\bbR(\gamma_p)) + \sum_{k = 0}^K\alpha_k\bbS^k
\end{equation}
where the $p$th approximated single pole filter $\bbH_T(\bbR(\gamma_p))$ is defined in \eqref{eq.FIRparam} and the parametric shift operator $\bbR(\gamma_p)$ in \eqref{eq.JacobiK}. In summary, a Jacobi approximation of the ARMA filter with orders $(P,T, K)$ is the one defined by \eqref{eq.FIRparam} and \eqref{eq.ARMAtot}. Scalar $P$ indicates the number of poles, $T$ the number of Jacobi iterations, and $K$ the order of the direct term $\sum_{k = 0}^K\alpha_k\bbS^k$ in \eqref{eq.PFD_ARMA}.

Substituting \eqref{eq.ARMAtot} into \eqref{eqn:graphPerceptron} yields an ARMA graph perceptron, which is the building block for ARMA GNNs or, for short, ARMANets. ARMANets are themselves convolutional. For a sufficiently large number of Jacobi iterations $T$, \eqref{eq.ARMAtot} is equivalent to \eqref{eq.ARMAfilt} which performs a pointwise multiplication in the spectral domain with the response \eqref{eq.ARMAResp}. The Jacobi filters in \eqref{eq.ARMAtot} are also reminiscent of the convolutional filters in \eqref{eqn:graphConv}. But the similarity is superficial because in ARMANets we train also the $2P$ single pole filter coefficients $\beta_p$ and $\gamma_p$ alongside the $K+1$ coefficients of the direct term $\sum_{k = 0}^K\alpha_k\bbS^K$. The equivalence suggests ARMANets may help achieve more discriminatory filters by tuning the single pole filter orders $P$ and $T$. An example of an implementation of ARMANets are CayleyNets \cite{Levie17-CayleyNets}, see \cite{Isufi20-EdgeNets}.

\subsection{EdgeNet}\label{subsec:EdgeNet}

While ARMANets enhance the discriminatory power of GCNNs with alternative convolutional filters, the edge varying GNN departs from the convolutional prior to improve GCNNs. The EdgeNet leverages the sparsity and locality of the shift operator $\bbS$ and forms a graph perceptron [cf. \eqref{eqn:GCNN}] by replacing the graph convolutional filter with an edge varying graph filter \cite{Coutino19-Distributed}.


\myparagraph{From shared to edge parameters.} In the convolutional filter \eqref{eqn:graphConv}, all nodes share the same scalar $h_k$ to weigh equally the information from all $k$-hop away neighbors $\bbS^k\bbx$. This is advantageous because it limits the number of trainable parameters, allows permutation equivariance, and favors stability. However, this parameter sharing limits also the discriminatory power to architectures whose filters $\bbH(\bbS)$ have the same eigenvectors as $\bbS$ [cf. \eqref{eqn:graphConvGFT}]. We can improve the discriminatory power by considering a linear filter in which node $i$ uses a scalar $\Phi_{ij}^{(k)}$ to weigh the information of its neighbor $j$ at iteration $k$. For $k = 0$, each node weighs only its own signal to build the zero-shifted signal $\bbz^{(0)} = \bbPhi^{(0)}\bbx$, where $\bbPhi^{(0)}$ is an $N \times N$ diagonal matrix of parameters with $i$th diagonal entry $\Phi_{ii}^{(0)}$ being the weight of node $i$. Signal $\bbz^{(0)}$ is subsequently exchanged with neighboring nodes to build the one-shifted signal $\bbz^{(1)} = \bbPhi^{(1)}\bbz^{(0)}$, where the parameter matrix $\bbPhi^{(1)}$ shares the support with $\bbI+\bbS$; the  $(i,j)$th entry $\Phi_{ij}^{(1)}$ is the weight node $i$ applies to signal $z_j^{(0)}$ from neighbor $j$. Repeating the latter for $k$ shifts, we get the recursion
\begin{equation}\label{eq.shiftedEV-signal}
    \bbz^{(k)} = \bbPhi^{(k)}\bbz^{(k-1)} = \prod_{k^\prime = 0}^k\bbPhi^{(k^\prime)}\bbx = \bbPhi^{(k:0)}\bbx,\quad k = 0, \ldots, K
\end{equation}
where the product matrix $\bbPhi^{(k:0)} = \prod_{k^\prime = 0}^k\bbPhi^{(k^\prime)} = \bbPhi^{(k)}\cdots\bbPhi^{(0)}$ accounts for the weighted propagation of the graph signal $\bbz^{(-1)} = \bbx$ from at most $k$-hops away neighbors. Each node is therefore free to adapt its weights for each iteration $k$ to capture the necessary local detail.

\myparagraph{Edge varying filters and EdgeNets.} The collection of signals $\bbz^{(k)}$ in \eqref{eq.shiftedEV-signal} behaves like a sequence of parametric shifts, where at iteration $k$ we use the parametric shift operator $\bbPhi^{(k)}$ to shift-and-weigh the signal. Following the same idea as in \eqref{eqn:graphConv}, we can sum up \emph{edge varying shifted} signals $\bbz^{(k)}$ to get the input-output map $\bby = \bbH(\bbPhi)\bbx$ of an edge varying graph filter. For this relation to hold, the filter matrix $\bbH(\bbPhi)$ should satisfy
\begin{equation}\label{eq.EdgeVarFilter}
    \bbH(\bbPhi) = \sum_{k = 0}^K\bbPhi^{(k:0)} = \sum_{k = 0}^K\bigg(\prod_{k^\prime = 0}^k\bbPhi^{k^\prime}\bigg).
\end{equation}
The edge varying graph filter is characterized by the $K+1$ parameter matrices $\bbPhi^{(0)}, \ldots, \bbPhi^{(K)}$ and contains $K(M+N)+N$ parameters. 
The edge varying graph filter forms the broadest family of graph filters: it generalizes the FIR filter in \eqref{eqn:graphConv} (for $\bbPhi^{(k:0)} = h_k\bbS^k$), the ARMA filter in \eqref{eq.ARMAtot}, and almost all other filters employed to build GNNs including spectral filters \cite{Bruna14-DeepSpectralNetworks}, Chebyshev filters \cite{Defferrard17-CNNGraphs}, Cayley filters, graph isomorphism filters, and also graph attention filters \cite{Velickovic18-GraphAttentionNetworks}.

Substituting \eqref{eq.EdgeVarFilter} into \eqref{eqn:graphPerceptron} yields an edge varying graph perceptron, which is the building block for edge varying GNNs or, for short, EdgeNets. EdgeNets are more than convolutional architectures; the high number of degrees of freedom and linear complexity render EdgeNets strong candidates for highly discriminatory GNNs in sparse graphs. If the graph is large, the EdgeNet can efficiently trade some edge detail (e.g., allowing edge varying weights only to a few nodes) to make the number of parameters independent of the graph dimension \cite{Isufi20-EdgeNets}. To control the number of parameters in EdgeNets we can adopt graph attention networks \cite{Velickovic18-GraphAttentionNetworks}, see \cite{Isufi20-EdgeNets} for details on this and other alternatives.


\section{Applications} \label{sec:applications}



We consider the application of GNNs for rating prediction in recommender systems (Section~\ref{subsec:recSys}) and learning decentralized controllers for flocking (Section~\ref{subsec:flocking}). These two applications aim at illustrating the use of GNNs in problems beyond semi-supervised learning.

We focus on the representation space of GNNs built with different filter \emph{types} and compare them with that of linear FIR filters to corroborate the discussed insights. We note that, in all cases, the values of hyperparameters (number of layers $L$, filter taps $K$ and features $F_{\ell}$) are design choices that have been made after cross-validation\footnote{The PyTorch GNN library used is available at \url{http://github.com/alelab-upenn/graph-neural-networks}.}.


\subsection{Recommender Systems} \label{subsec:recSys}

Consider the problem of rating prediction in recommender systems. We have a database of users that have rated many items, and we use it to build a graph where each item is a node and each edge weight is given by the rating similarity between items \cite{Huang18-RatingGSP}. Then, given the ratings a specific user has given to some of the items, we want to predict the rating the same user would give to a specific item not yet rated. The ratings given by that user can be modeled as a graph signal, so that this becomes a problem of interpolating one of the (unknown) entries in it.

\myparagraph{Setup.} We consider items as movies and use a subset of the MovieLens-100k dataset, containing the $200$ movies with the largest number of ratings \cite{Harper16-MovieLens}. The resulting dataset has $47,825$ ratings given by $943$ users to some of those $200$ movies. The similarity between movies is the Pearson correlation \cite[eq. (6)]{Huang18-RatingGSP}, which is further sparsified to keep only the ten edges with the stronger similarity. We split the dataset into $90\%$ for training and $10\%$ for testing. In this context, each user represents a graph signal, where the value at each node is the rating given to that movie. Movies not rated are given a value of zero. The objective is to estimate the rating a user would give to the movie \emph{Star Wars} based on the ratings given by that same user to other movies and leveraging the graph of rating similarities.

\myparagraph{GNN models and training.} We implement a FIR graph filter \eqref{eqn:singleFilter}-\eqref{eqn:singleTensorConv} [cf. \cite{Huang18-RatingGSP}], a GCNN \eqref{eqn:singleFilter}-\eqref{eqn:singleTensorConv}-\eqref{eqn:singleNonlinearity}, an ARMANet with $T = 1$ Jacobi iterations \eqref{eq.ARMAtot}, and an EdgeNet \eqref{eq.EdgeVarFilter}. The number of features in all cases is $F_{1}=64$, the filter order is $K=4$, and ReLU nonlinearities are used. We include a local readout layer and extract the entry corresponding to the \emph{Star Wars} movie as the estimate of the rating. The loss function is the smooth L1 loss $J(x,y) = 0.5(x-y)^{2}$ if $|x-y| < 1$ and $|x-y|-0.5$ otherwise, and the evaluation measure is the root MSE (RMSE). We train the architectures for $40$ epochs with a batch size of $5$ samples, using ADAM \cite{Kingma15-ADAM} with learning rate $5 \times 10^{-3}$ and forgetting factors $0.9$ and $0.999$.

%
\begin{figure}
    \centering
    \includegraphics[width=0.8\textwidth]{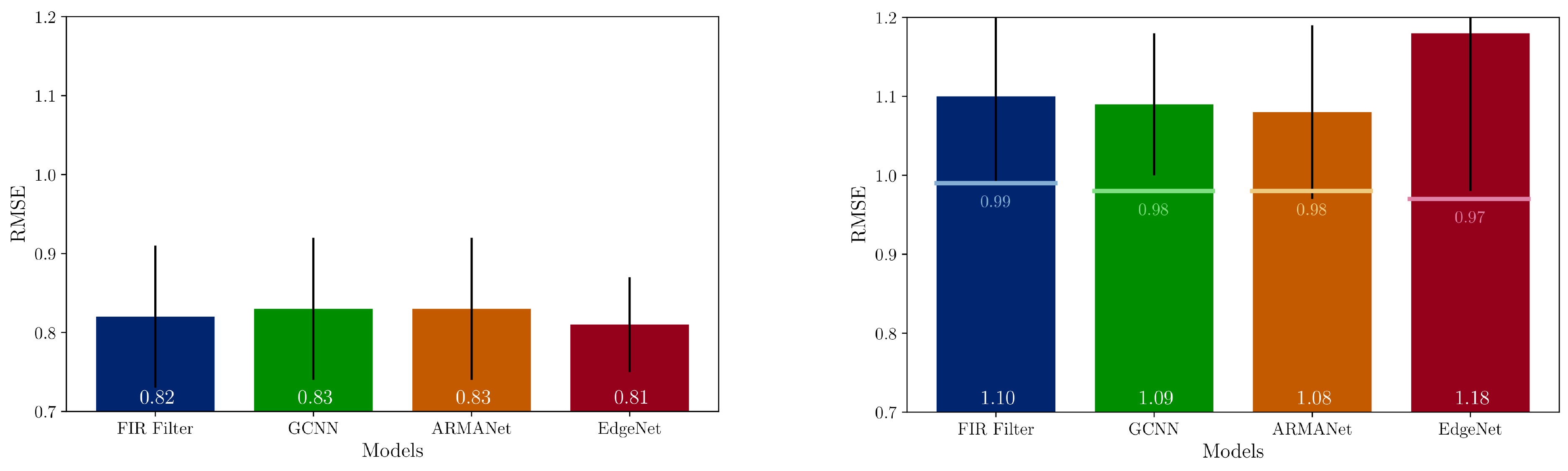}
    \caption{RMSE and standard deviation for different architectures in the movie recommendation problem. (Left) Training and testing on \emph{Star Wars}. (Right) Training on \emph{Star Wars}, testing on \emph{Contact}; we also include the RMSE obtained for training and testing on the movie \emph{Contact} as horizontal solid lines.}
    \label{fig:recSys}
\end{figure}
%

\myparagraph{Results.} Fig.~\ref{fig:recSys} (left) shows that the rating for all models is similar, with the EdgeNet performing slightly better at an RMSE of $0.81 (\pm0.05)$. In Fig.~\ref{fig:recSys} (right) we take the same models trained for estimating the rating for \emph{Star Wars}, and extract the rating predicted for \emph{Contact} instead. We do this in an attempt to show transferability of the trained models. In this case, the performance is similar among the FIR filter, GCNN and ARMANet, but the EdgeNet has severely degraded.

\myparagraph{Discussion.} First, we note that the GCNN performs similarly to the linear graph filter which, in light of Theorem \ref{thm:stability}, suggests that the relevant content is in low eigenvalues. Second, the GCNN and the ARMANet exhibit essentially the same performance, suggesting that the ARMA filter does not significantly increase the representation power, which is true in light of the FIR implementation (Jacobi) of ARMA filters. Third, the EdgeNet achieves the best performance when training and testing for the same movie, suggesting an increase in representation power, but does not transfer well to other settings, likely because it does not satisfy Theorems \ref{thm:permutationEquivariance} nor \ref{thm:stability}.


\subsection{Learning Decentralized Controllers for Flocking} \label{subsec:flocking}

The objective of flocking is to coordinate a team of agents to fly together with the same velocity while avoiding collisions. Agents start flying at arbitrary velocities and need to take appropriate actions to flock together. This problem has a straightforward \emph{centralized} solution that amounts to setting each agent's velocity to the average velocity of the team \cite[eq. (10)]{Tolstaya19-Flocking}, but \emph{decentralized} solutions are famously difficult to find \cite{Witsenhausen68-Counterexample}. Since GCNNs are naturally distributed, we use them to \emph{learn} the decentralized controllers.

\begin{figure}[t]
    \centering
    \includegraphics[width=0.85\textwidth]{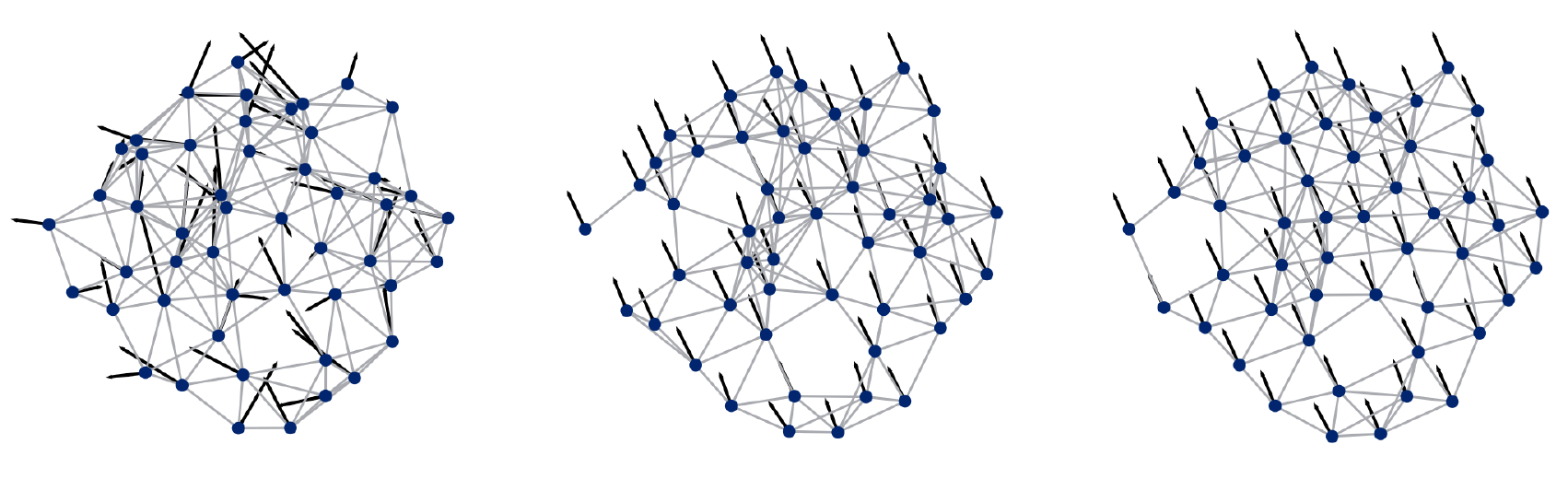}
    \caption{Snapshots of a sample trajectory. The dots illustrate the agents, the gray edges represent the communication links, and the arrows show the velocity. (Left) The agents start flying at time $t=0\text{s}$ with arbitrary velocities. (Middle) They manage to agree on a direction at $t=1\text{s}$. (End) And they effectively fly together at $t=2\text{s}$.}
    \label{fig:trajectory}
\end{figure}

\myparagraph{Setup.} Let us consider a team of $N=50$ agents, in which each agent $i$ is described at discrete time $t$ by its position $\bbr_{i}(t) \in \reals^{2}$, velocity $\bbv_{i}(t) \in \reals^{2}$, and acceleration $\bbu_{i}(t) \in \reals^{2}$. We want to control the acceleration $\bbu_{i}(t)$ of each agent, so that they coordinate their velocities $\bbv_{i}(t)$ to be the same for all $i$, see Fig.~\ref{fig:trajectory}. We consider a decentralized setting where agents $i$ and $j$ can communicate with each other at time $t$ only if $\|\bbr_{i}(t) - \bbr_{j}(t)\| \leq R = 2\text{m}$. This defines a communication graph $\ccalG(t)$ that changes with $t$ as the agents move around, imposing a delayed information structure \cite[eq. (2)]{Tolstaya19-Flocking}. Note that we can easily adapt graph filters (and thus, GCNNs) to the change in support matrices $\bbS(t)$ by using delayed FIR filters $\bbH(\bbS(t))\bbx(t) = \sum_{k=0}^{K} h_{k} \bbS(t) \bbS(t-1) \cdots \bbS(t-k+1) \bbx(t-k)$ [cf. \eqref{eqn:graphConv}]. The filter taps $h_{k}$ are the same, but the shift operators change with time. We generate $400$ optimal trajectories for training and $20$ for testing. We refer to \cite{Tolstaya19-Flocking} for details on system dynamics.

\myparagraph{GNN models and training.} We implement a linear FIR filter and a GCNN. We consider $F_{1}=32$ features and filters of order $K_{1} = 3$. We include a second, local readout layer to obtain the final acceleration $\bbu_{i}(t) \in \reals^{2}$ that each agent takes. We train the architectures using imitation learning by minimizing the MSE between the output action $\bbu_{i}(t)$ and the optimal action $\bbu_{i}^{\star}(t)$ given by \cite[eq. (10)]{Tolstaya19-Flocking}. At test time, we do not require access to the optimal action. The evaluation measure is the velocity variation of the team throughout the trajectory, $N^{-1} \sum_{t} \sum_{i=1}^{N} \| \bbv_{i}(t) - \barbv(t)\|^{2}$ with $\barbv(t) = N^{-1} \sum_{j=1}^{N} \bbv_{j}(t)$ being the average velocity at time $t$. We trained for $40$ epochs with a batch size of $20$ samples using ADAM \cite{Kingma15-ADAM} with learning rate $5\times 10^{-4}$ and forgetting factors $0.9$ and $0.999$.

%
\begin{table}
    \centering
    \caption{Scalability. Trained on $50$ agents. Tested on $N$ agents. Optimal cost: $51 (\pm 1)$.}
    {\footnotesize
    \begin{tabular}{|c|c|c|c|c|c|} \hline
        $N$ & $50$ & $62$ & $75$ & $87$ & $100$ \\ \hline
        FIR filter & $408(\pm88)$ & $408(\pm93)$ & $434(\pm128)$ &  $420(\pm105)$ & $430(\pm131)$\\ \hline
        GCNN & $77 (\pm 3)$ & $78 (\pm 3)$ & $77 (\pm 2)$ & $77 (\pm 2)$ & $78 (\pm 2)$ \\ \hline
    \end{tabular}
    }
    \label{tab:flockingStable}
\end{table}
%

\myparagraph{Results.} We observe in Table~\ref{tab:flockingStable} the cost achieved by the GCNN-based controller is close to the optimal cost, while the FIR filter fails to control the system leading to a very high cost. We further investigate the effect of Theorems \ref{thm:permutationEquivariance}~and~\ref{thm:stability} by transferring at scale the learned solutions. That is, we take the controllers learned with teams of $N=50$ agents, and test them in teams of increasing size. The GCNN scales perfectly, maintaining the same performance.

\myparagraph{Discussions.} The GCNNs improved performance over the graph filter is expected since we know that optimal distributed controllers are nonlinear \cite{Witsenhausen68-Counterexample}. The GCNN also achieves a cost close to optimum, evidencing successful control. Once trained, this GCNN based controller can be used in teams of arbitrary number of agents evidencing the properties of permutation equivariance and stability, and speaking to the potential of GCNNs for learning behaviors in homogenous teams.


\section{Conclusion} \label{sec:conclusion}



Graph signal processing plays a crucial role in characterizing and understanding the representation space of graph neural networks. By emphasizing the role of graph filters and leveraging the concept of graph Fourier transform, we are able to derive fundamental properties such as permutation equivariance and stability, as well as establish a unified mathematical description. This reinforces the notion that GNNs are nonlinear extensions of graph filters, and thus GSP can help explain and understand the observed success of GNNs and contribute to improved designs.

As a matter of fact, several areas of interest lie ahead for GSP researchers to pursue. First, the understanding of what precise effect the nonlinearities have on the frequency content is limited. A better characterization of their effect in relation to the underlying topology is bound to help in designing appropriate ones. Second, the general relationship between the hyperparameters (number of layers, filter taps) and the characteristics of the graph (diameter, degree) is currently unknown. It is expected, for instance, the number of hops bears some relationship with the diameter of the graph, but no theoretical result is out there yet. Third, the bounds in the stability results are quite loose due to the coarse bound used on the eigenvectors. Thus, focusing on the eigenvector perturbation to improve the bound is a worthwhile pursuit. Fourth, the stability result holds for graphs of the same size. Extending this result to graphs of different size is an important research direction. Finally, we mention exploring the possibility of nonlinear aggregations of the filter banks, as well as using different shift operators at each layer.

From a higher vantage point, realizing GNNs are an object of study of GSP and regarding them as nonlinear extensions of graph filters, help us exploit our understanding of filtering techniques as well as leverage spectral domain analysis. Thus, GSP plays a crucial role in characterizing, understanding, and improving GNNs.

%
\urlstyle{same}
\bibliographystyle{IEEEtran}
\bibliography{biblio/myIEEEabrv,biblio/biblioGSP}

\end{document}